\def\year{2015}
\documentclass[letterpaper]{article}
\usepackage{aaai}
\usepackage{times}
\usepackage{helvet}
\usepackage{courier}

\usepackage{multirow}
\usepackage{epstopdf}
\usepackage{pgfplots}
\usepackage{algorithmic}
\usepackage{algorithm}
\usepackage{subfigure}
\usepackage{amsmath}
\usepackage{amssymb}  % assumes amsmath package installed
\usepackage{booktabs}
\usepackage{tabularx}

\begin{document}
% The file aaai.sty is the style file for AAAI Press 
% proceedings, working notes, and technical reports.
%
\title{HCLAE: High Capacity Locally Aggregating Encodings for Approximate Nearest Neighbor Search}
\author{
  Liu Shicong, Shao Junru, Lu Hongtao\\
  \texttt{\{artheru, yz\_sjr, htlu\}@sjtu.edu.cn Shanghai Jiaotong University}
}
\maketitle
\begin{abstract}
\begin{quote}
Vector quantization-based approaches are successful to solve Approximate Nearest Neighbor (ANN) problems which are critical to many applications. The idea is to generate effective encodings to allow fast distance approximation. We propose quantization-based methods should partition the data space finely and exhibit locality of the dataset to allow efficient non-exhaustive search. In this paper, we introduce the concept of High Capacity Locality Aggregating Encodings (HCLAE) to this end, and propose Dictionary Annealing (DA) to learn HCLAE by a simulated annealing procedure. The quantization error is lower than other state-of-the-art. The algorithms of DA can be easily extended to an online learning scheme, allowing effective handle of large scale data. Further, we propose Aggregating-Tree (A-Tree), a non-exhaustive search method using HCLAE to perform efficient ANN-Search. A-Tree achieves magnitudes of speed-up on ANN-Search tasks, compared to the state-of-the-art.
\end{quote}
\end{abstract}

\section{Introduction}
Approximate nearest neighbor (ANN) search is a fundamental problem in many computer science topics, especially in those involving high-dimensional and large-scale datasets like machine learning, pattern recognition, computer vision, information retrieval, etc, due to the high computation efficiency requirements. Among existing ANN techniques, quantization-based algorithms(\cite{pq},\cite{opq},\cite{composite}, etc.) have shown the state-of-the-art performances by allowing efficient distance computation via asymmetric distance computation (ADC)\cite{pq} between a query vector and an encoded vector. One can perform an exhaustive ADC to retrieve the approximate nearest neighbor.

Even so, an exhaustive comparison between the query and the dataset is still prohibitive for even larger datasets like \cite{80m}. IVFADC \cite{pq} provides non-exhaustive search based on coarse quantizers and encoded residues. The idea is to obtain a candidates list possibly containing the nearest neighbor, then perform ADC on the list. Similar methods like inverted multi-index\cite{babenko2012inverted}, Locally Optimized Product Quantization\cite{kalantidis2014locally}, Joint Inverted Indexing \cite{xia2013joint}, etc, has various improvements.

\subsubsection{Problems of existing quantization-based algorithms.}
One challenge in designing non-exhaustive search algorithm is: the locality of a vector is not exhibited in the encoding. Thus, researchers have to do some roundabout to dig out the locality, like using a coarse quantizer. These methods lack efficiency because candidate listing and re-ranking are totally irrelevant. In addition, we would like the encodings to have high capacity w.r.t the data space, i.e. to distinguish more vectors, so the data space can be effectively represented. However, existing quantization methods didn't explicitly consider these issues.

\subsubsection{Major Contributions}In this paper, we are interested in encodings which not only accelerate distance computation, but also 'aggregate' the locality of a dataset, along with high capacities. We introduce the concept of High Capacity Locally Aggregating Encodings (HCLAE) for ANN-search to address the aforementioned problems. We propose \emph{Dictionary Annealing} (DA) algorithm to generate HCLAE encodings of the dataset. Inspired by simulated annealing, the main idea of DA is to "heat up" a dictionary with current residue, then "cool down" the dictionary to reduce the residue. Auxiliary algorithms for DA are also introduced to further increase capacity and to reduce distortion. DA is naturally an online learning algorithm and is suitable for large scale learning.

To utilize HCLAE encodings on large scale data, we propose \emph{Aggregating Tree} (A-Tree) for fast non-exhaustive search. It's a radix-tree like structure based on the encoding of the dataset, so the common prefixes of the encodings can be effectively represented with one node. A-tree is memory efficient and allows fast non-exhaustive search: we breadth first traverse the tree with a priority queue to obtain the candidate list. The time consumption is significantly lower than other non-exhaustive search methods.

We have validated DA and A-Tree on various standard benchmarks: SIFT-1M, GIST-1M\cite{pq}, SIFT-1B\cite{jegou2011searching}. Empirical Results show DA improves the quantization of dataset greatly, and A-Tree can bring magnitudes of speed up compared to existing non-exhaustive search methods. The overall performance of DA and A-Tree outperforms existing state-of-the-art methods. The online version also shows great practical interest. Applications depending on ANN search cam greatly benefit from our algorithms.

\section{Background and Motivation}

The main idea of quantization-based methods is to generate encodings consisting of $M$ parts for fast distance computation. For example, Product Quantization\cite{pq} splits the data space into $M$ disjoint subspaces, and separately learns dictionaries for each subspaces, then quantizes each subspace to produces encodings of a vector $\mathbf{x} \rightarrow \{i_1(\mathbf{x}), i_2(\mathbf{x}), \cdots, i_M(\mathbf{x}) \}$. PQ allows fast approximate distance computation between a query vector and an encoded vector via Asymmetric Distance Computation(ADC), which is discussed in detail in \cite{pq}, \cite{babenko2015tree}, \cite{babenko2014additive}.

However, in the real applications involving large scale data, exhaustively computing distances doesn't meet the query speed requirement. It's practical to perform some preprocessing such as candidates listing. IVFADC\cite{pq}, the Inverted Multi-index\cite{babenko2012inverted} and Locally Optimized Product Quantization\cite{kalantidis2014locally}, etc. are proposed to perform these tasks. However, these candidates listing methods are totally irrelevant to the encodings of the dataset, adding additional computation and storage cost.

\begin{figure}[t]
\centering 
\subfigure[AQ]{
\includegraphics[width=0.22\linewidth]{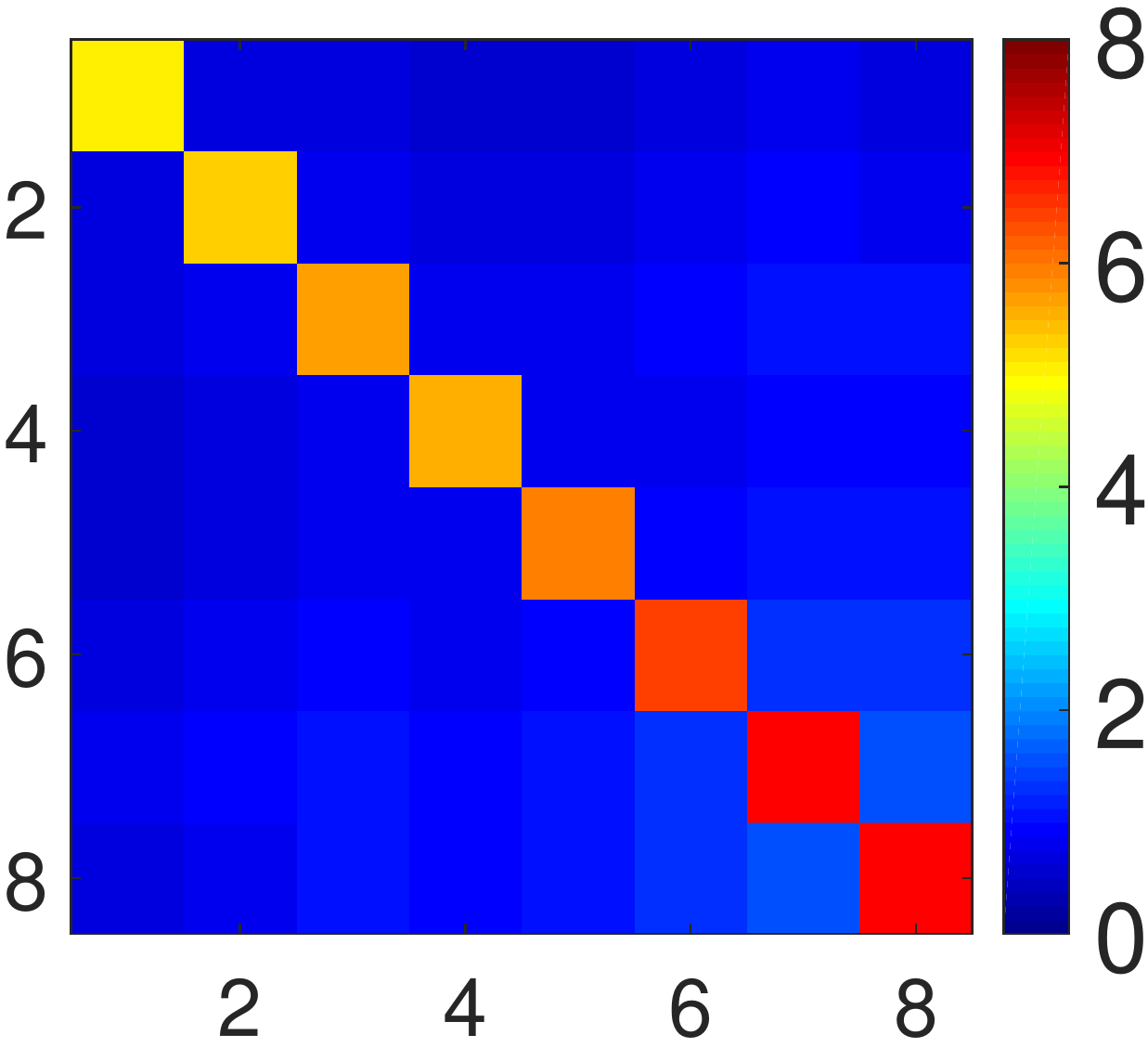}
}
\subfigure[OPQ]{
\includegraphics[width=0.22\linewidth]{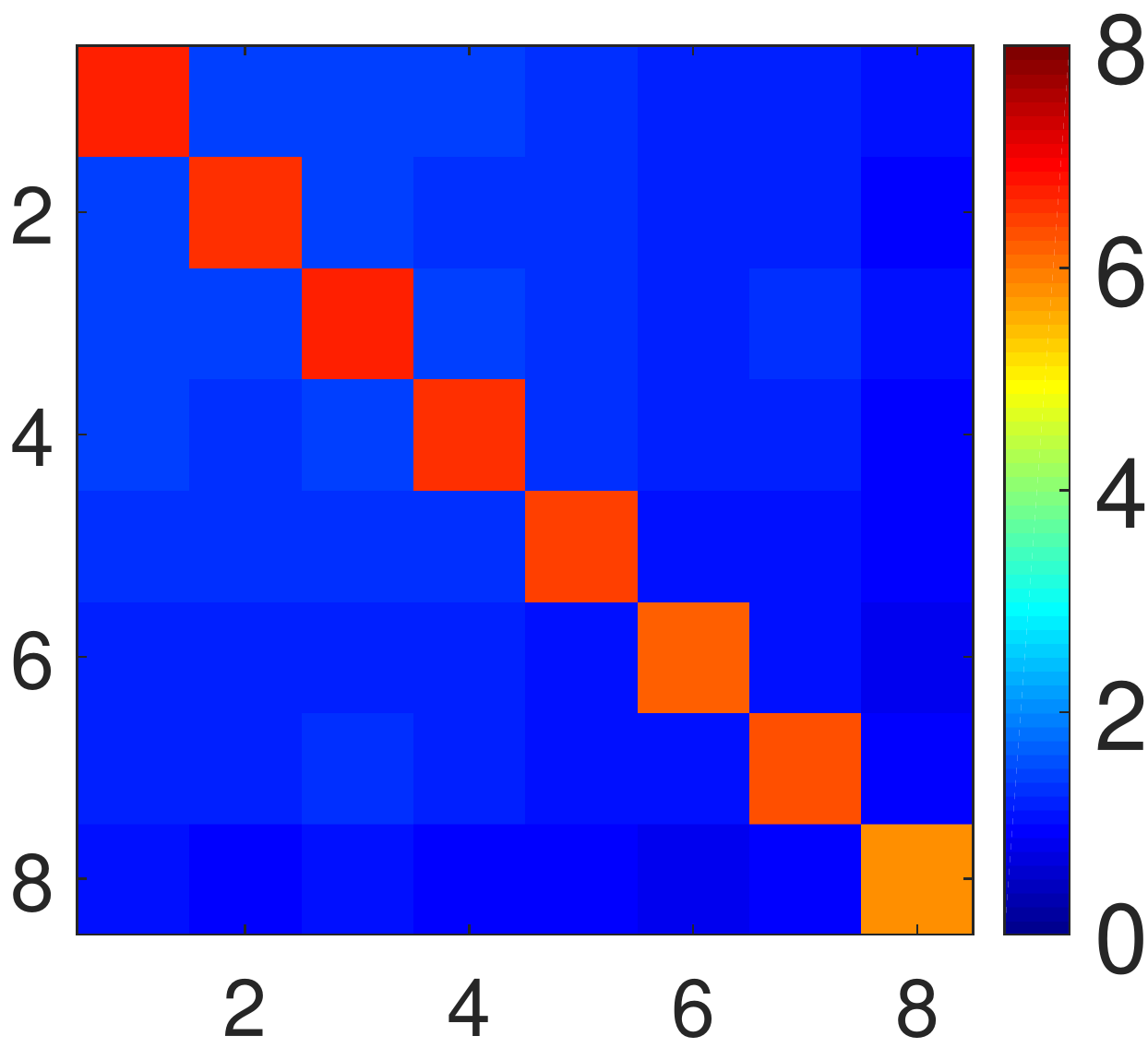}
}
\subfigure[RVQ]{
\includegraphics[width=0.22\linewidth]{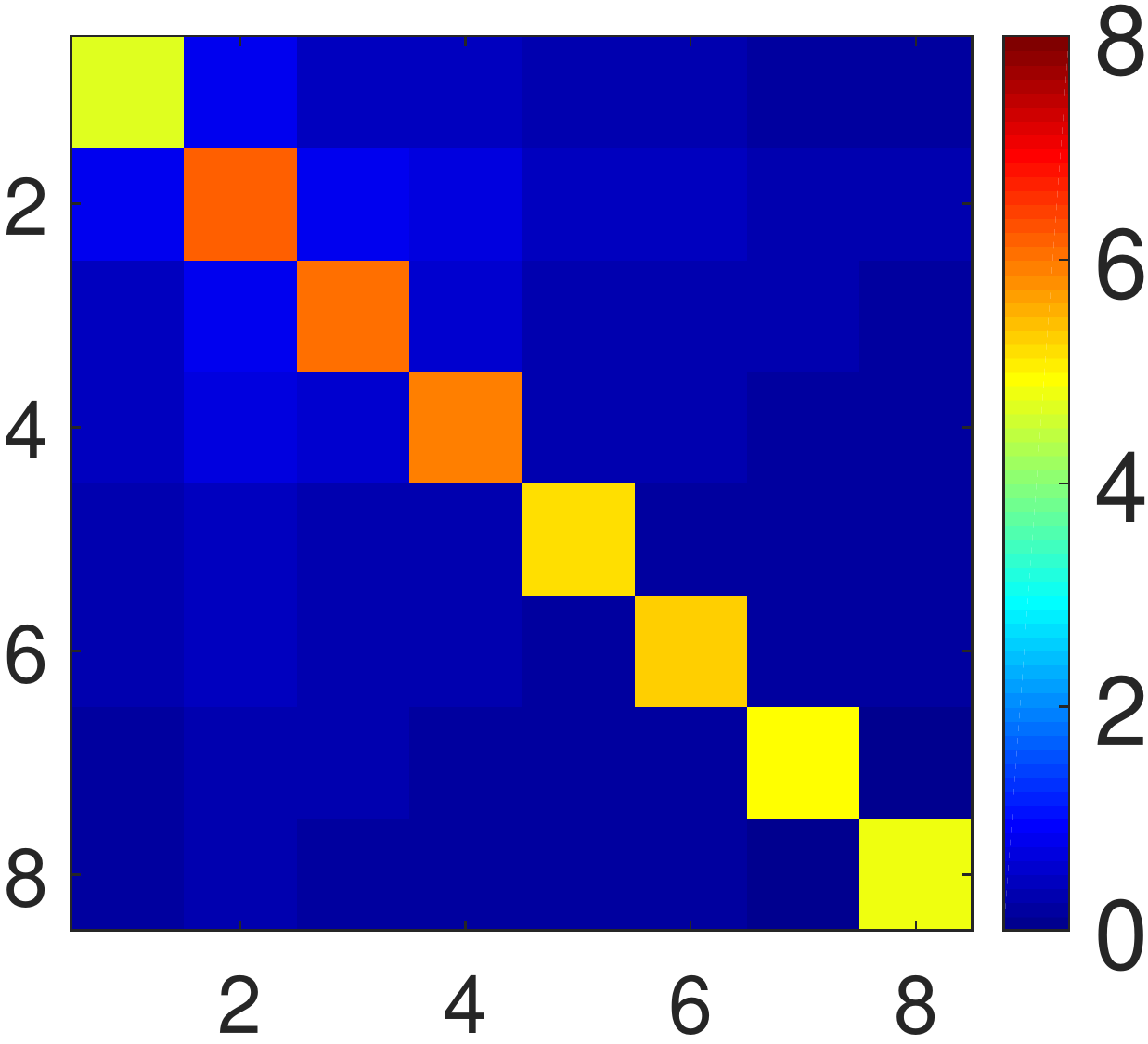}
}
\subfigure[\textbf{DA}]{
\includegraphics[width=0.22\linewidth]{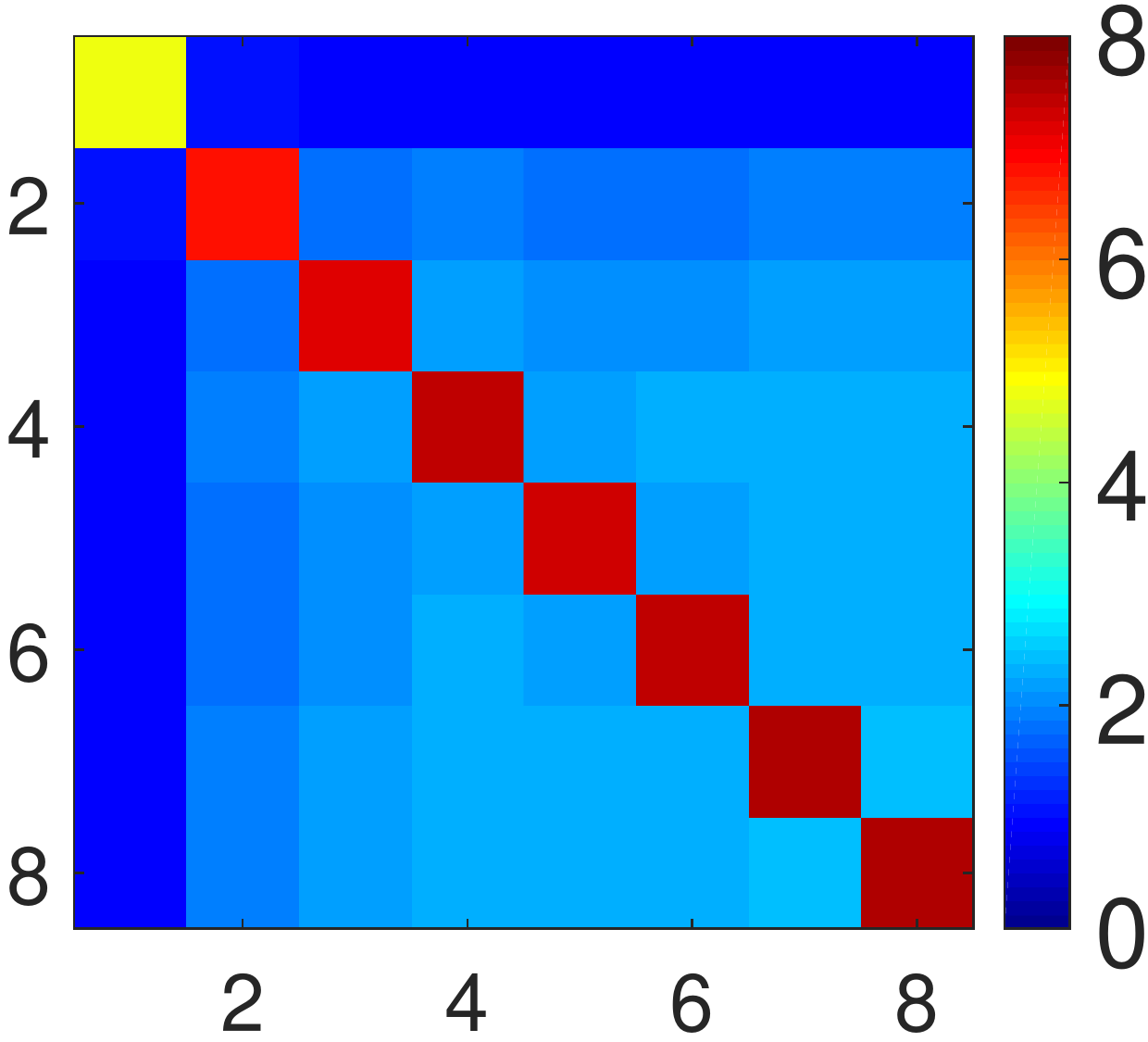}
}
\caption{Mutual information matrices of local vectors' encodings on GIST1M dataset, $M=8, K=256$, indicating Locally Aggregating of different methods.}
\label{MIL}
\end{figure}

\begin{figure}[t]
\centering 
\subfigure[AQ]{
\includegraphics[width=0.22\linewidth]{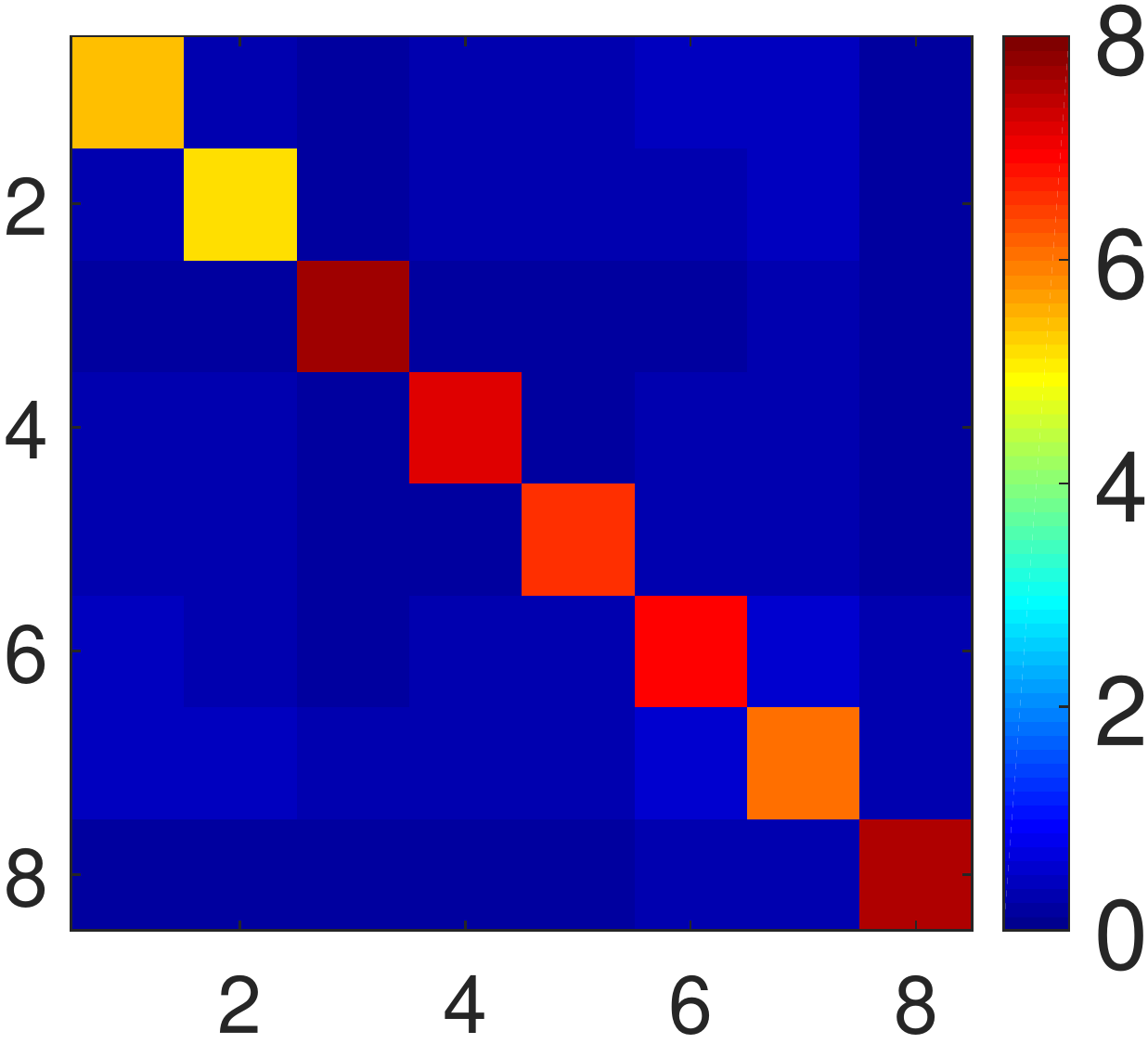}
}
\subfigure[OPQ]{
\includegraphics[width=0.22\linewidth]{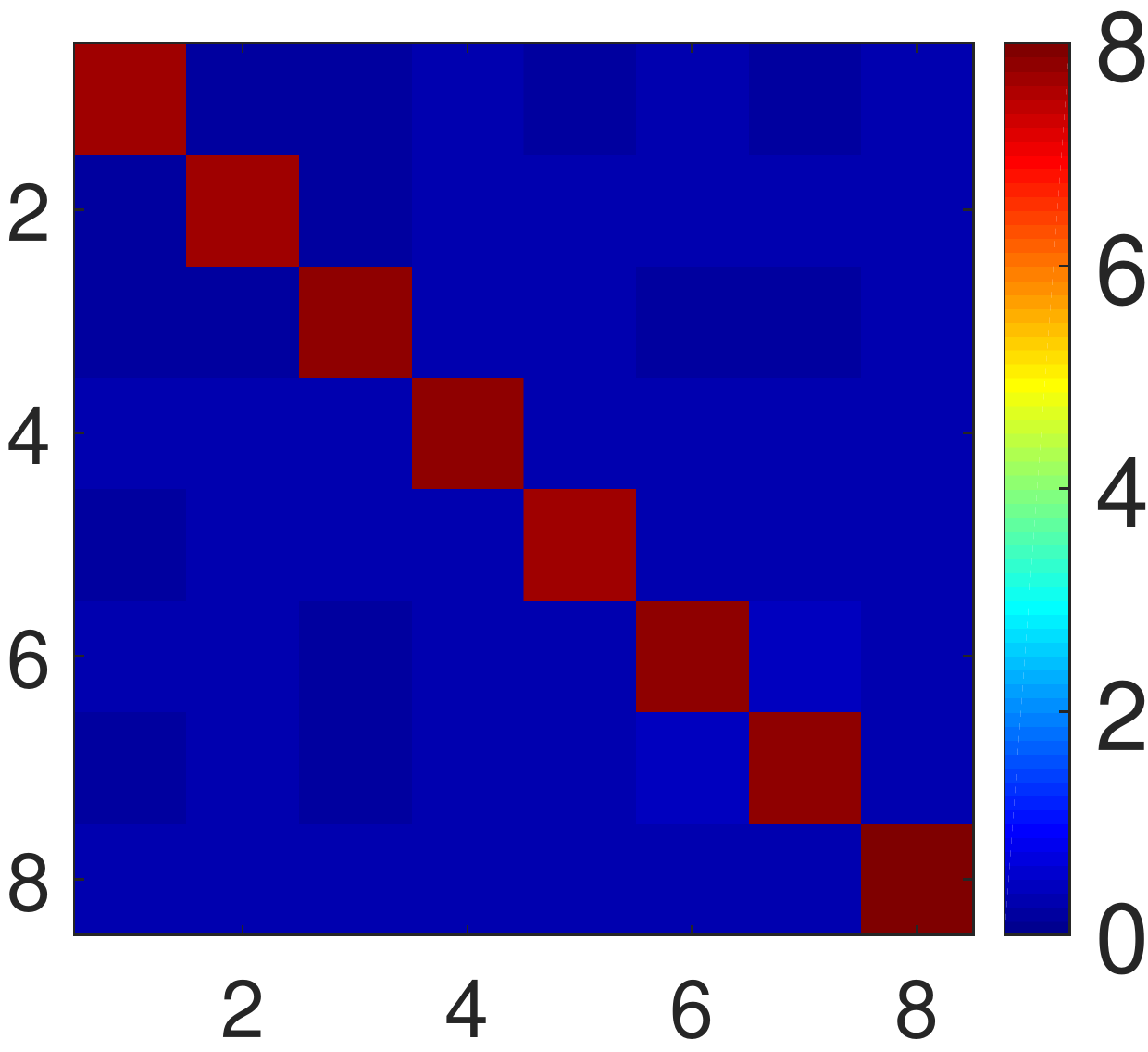}
}
\subfigure[RVQ]{
\includegraphics[width=0.22\linewidth]{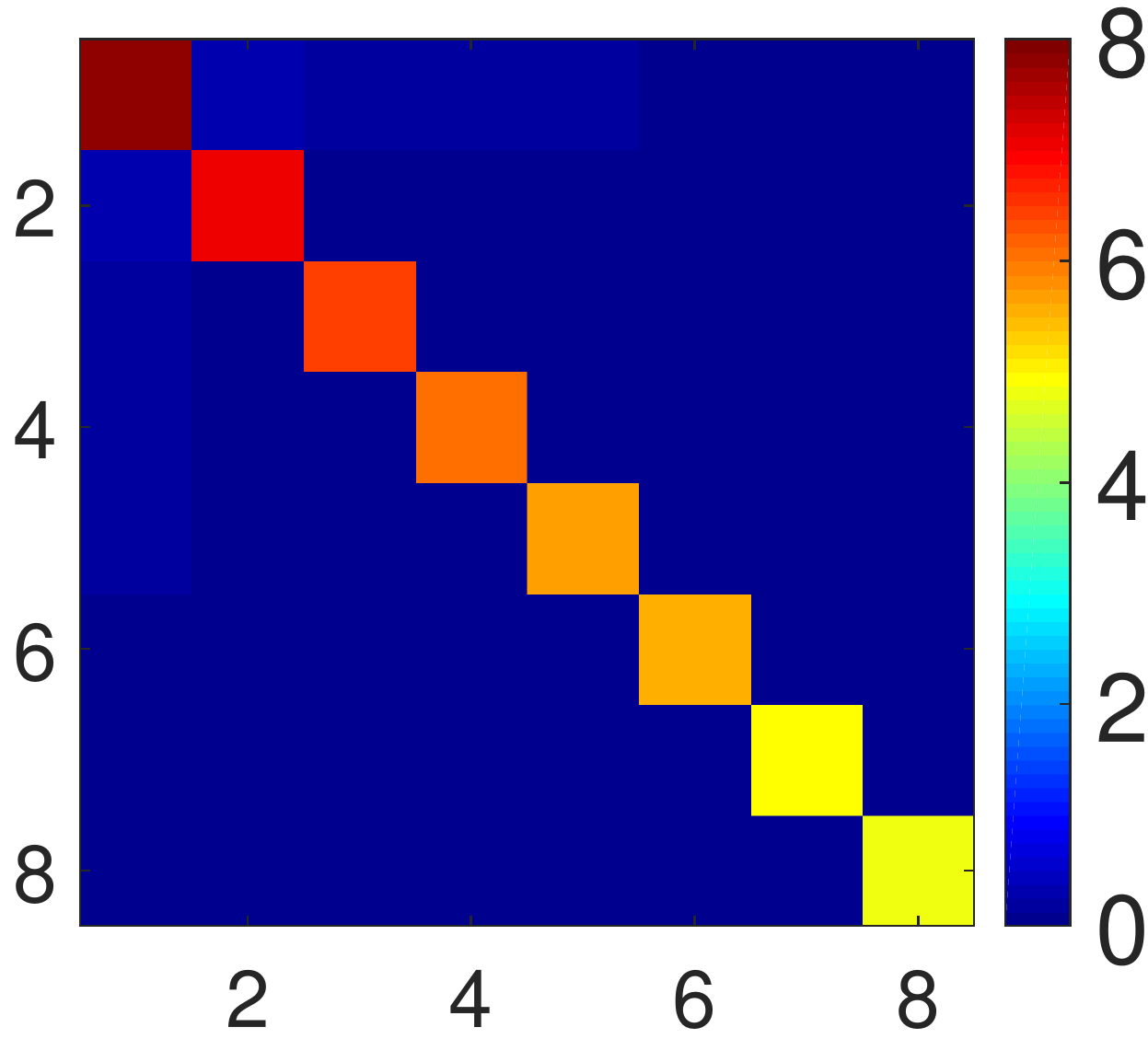}
}
\subfigure[\textbf{DA}]{
\includegraphics[width=0.22\linewidth]{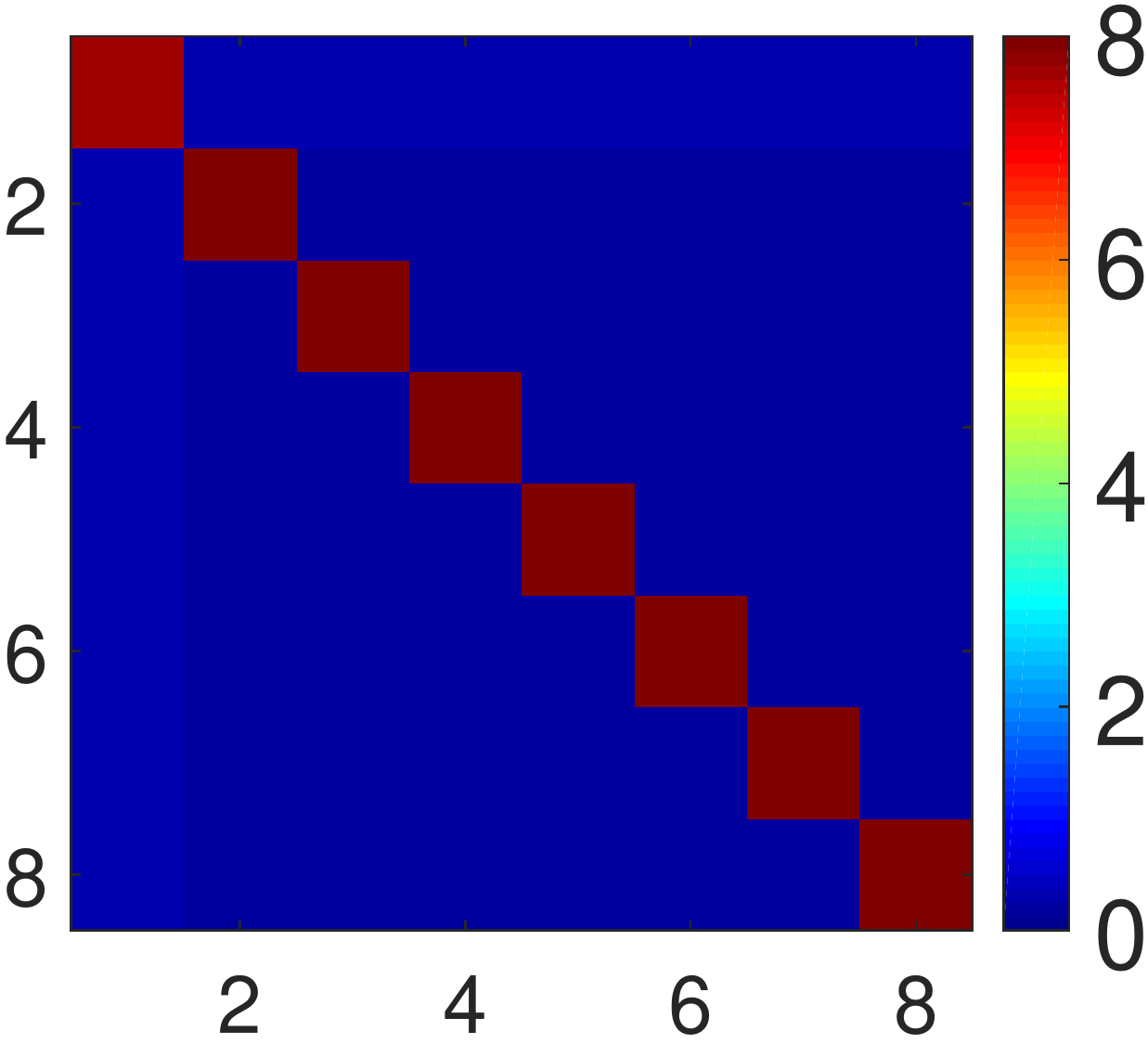}
}
\caption{Mutual information matrices of all encoded vectors on GIST1M dataset, $M=8, K=256$ indicating Encoding Capacity of different methods. Best viewed in color.}
\label{MI}
\end{figure}

\subsection{Locally Aggregating Encodings}
A common methodology for non-exhaustive search is bound-and-branch with trees. The effectiveness of tree structures lies in how can it effectively tell which child node contains the nearest neighbor. However, in high-dimensional space, tree structures like KD-Tree\cite{kd} generally degrades to linear scan because the nearest neighbor may be contained in \textit{any} node\cite{kdbad}. To utilize bound-and-branch methodology, this search scope must be able to narrow down.

Our solution is to utilize the priors of the visited node: if a node is deep in the tree, then we know which child node may contain the nearest neighbor. We name it Locally Aggregating. Note one can transform encodings to a radix tree. Denote the $m$-th part encoding of a vector $\mathbf{x}$'s local vector $\mathbf{x'}$ as $I_m=i_m(\mathbf{x'})$, and $\mathcal{L}_m$ as the conditional entropy:
$$\mathcal{L}_m=H(I_m|I_1,I_2,\cdots,I_{m-1})$$
$\mathcal{L}_m$ directly measures to what extent can we narrow down the search scope, so a fast descending $\mathcal{L}_m$ is preferred. Directly computing $\mathcal{L}_m$ is not easy, nevertheless, we present the mutual information matrix of $I_m$ obtained with different quantization methods in Figure \ref{MIL} for visualization.

\subsection{Encoding Capacity}
To effective encode a dataset, we would like the data-space is partitioned finely, so vectors could be easily distinguished. It's straightforward to define the Encoding Capacity as the total information entropy: $\mathcal{S}=H(I_1,I_2,\cdots,I_M)$. In practice, optimizing encoding capacity is usually relaxed into two separate objectives:
\begin{enumerate}
\item Maximize self-information $H(I_m)$ for $m=1\cdots M$
\item Minimize mutual information $H(I_i; I_j)$ for $i,j=1\cdots M,i\ne j$
\end{enumerate}
The above objectives were explicitly considered in hashing methods including Spectral Hashing\cite{SH}, Semi-supervised Hashing\cite{wang2010semi}, etc. which are proposed to learn balanced and uncorrelated bits. For quantization methods, encoding capacity has not been addressed yet. In Figure \ref{MI}, we visualize the comparison of the encoding capacities of different quantization methods in mutual information matrix.

\section{Learning High Capacity Locally Aggregating Encodings (HCLAE)}
As described above, for a high capacity encoding, $H(\mathbf{I}_m)$ is maximized. By chain rule, to lower $\mathcal{L}_m=H(\mathbf{I}_1,\cdots,\mathbf{I}_m)-H(\mathbf{I}_m)$, $H(\mathbf{I}_1,\cdots,\mathbf{I}_m)$ should be minimized, I.e. the local vectors should have the same prefix encoding. By Lloyd's condition\cite{vec}, we could perform Residual Vector Quantization(RVQ)\cite{juang1982multiple}\cite{rvq} on the dataset. However for high dimensional data, the encoding capacity is low with RVQ and doesn't exhibit locally aggregating. We introduce Dictionary Annealing to produce High Capacity Locally Aggregating Encodings.

\subsection{Dictionary Annealing}

\begin{figure}
\begin{center}
%\fbox{\rule{0pt}{2in} \rule{0.9\linewidth}{0pt}}
%\includeCroppedPdf[width=\textwidth]{figures/test.eps}
%\includegraphics[width=0.8\linewidth, trim 35mm 100mm 40mm 40mm, clip]{figures/DA-illustration.pdf}
\includegraphics[width=1\linewidth]{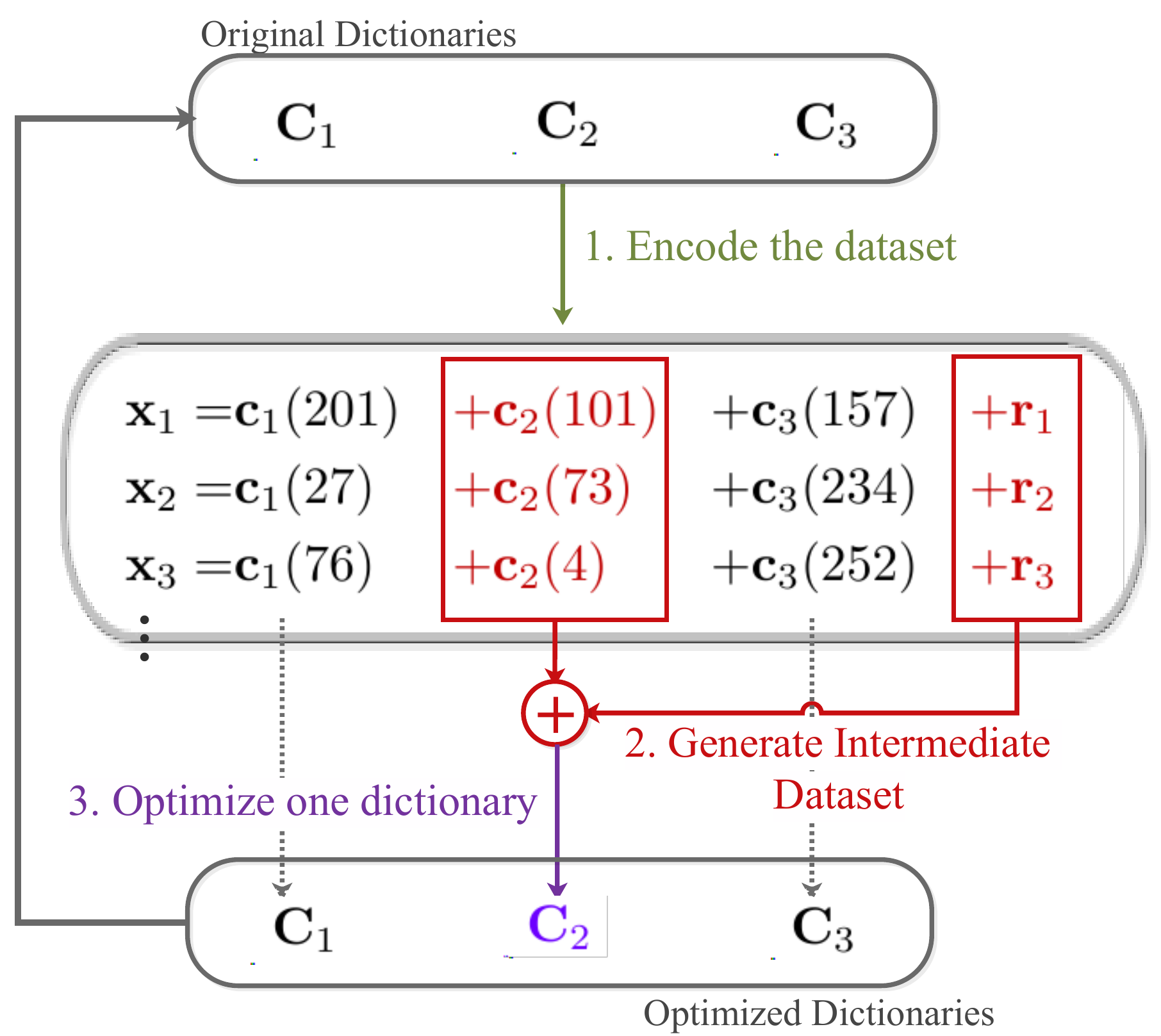}
\end{center}
   \caption{The illustration of one iteration of \textbf{Dictionary Annealing}. We perform several iterations to optimize the whole series of dictionaries. }
\label{figIllustration}
\end{figure}

Dictionary Annealing(DA) performs simulated annealing on an series of existing dictionaries, while it can also learn dictionaries from scratch. Figure \ref{figIllustration} provides an intuitive illustration of DA. To optimize a single dictionary $\mathbf{C}_m=\{\mathbf{c}_m(1), \cdots, \mathbf{c}_m(K) \}$, let's assume it's already at the local lowest energy position, i.e, not improving on the previous optimization/learning. We first "heat up" the dictionary, by putting the "noisy" residue $\{\mathbf{e}_\mathbf{x}\}$ into $\mathbf{C}_m$ to generate an intermediate dataset: $$\mathbf{x}'=\mathbf{e}_\mathbf{x}+\mathbf{c}_m(i_m(\mathbf{x}))$$,
 Then we "cool down" dictionary $\mathbf{C}_m$ by incrementally fitting $\{\mathbf{x}'\}$. 

Why the intermediate dataset and why using residues? We have two reasons:
\begin{itemize}
\item The intermediate dataset is the residue dropping currently optimizing dictionary. The quantization error is reduced if the intermediate dataset is better fitted. On the whole picture, this $m$-th dictionary does a better job and residues left for the next dictionary is lowered, lowering $H(\mathbf{I}_{m+1}|\mathbf{I}_1,\cdots,\mathbf{I}_m)$.

\item The residues are independent to other dictionary spaces, as they're "noises" to these dictionaries. Messing with residues won't rise mutual information between dictionaries. So we can push the $H(\mathbf{I}_m)$ higher without worry.
\end{itemize}

Given a series of dictionaries, the algorithm is performed by multiple iterations. On each iteration, we optimize one dictionary, then re-encode the dataset to obtain the new residue for the next iteration. To learn dictionaries from scratch, one can simply perform DA on "all-zeros" dictionaries, \footnote{In this case DA is quite similar to Residual Vector Quantization: the intermediate dataset of an "all-zeros" dictionary is the same to the residues}. To bring better performance, we propose the following two auxiliary algorithms:

\subsection{Improved K-means for High-dimensional Residue}
Clustering on high dimensional space is not easy, especially on high-dimensional residues as the randomness is increased. To obtain a better clustering for high dimensional data, one approach is to cluster on lower-dimensional subspace\cite{agrawal1998automatic}, which is also done by PQ/OPQ to obtain high information entropy for each dictionary. \cite{ding2004k} indicates that PCA dimension reduction is particularly beneficial for K-means clustering, as it finds best low rank L2 approximations. In addition, the dictionary learned previously can provide initial points good enough, which is important for k-means clustering\cite{bradley1998refining}.

Our idea is to preserve the clustering information on lower-dimensional subspace for higher-dimensional subspace clustering. To optimize dictionary $\mathbf{C}_m$ for $\{\mathbf{x'}\}$, we first designate a dimension adding sequence: $d_1 < d_2 < \cdots < d_I = d$, then:

\begin{enumerate}
\item Project $\mathbf{C}_m$ and $\{\mathbf{x'}\}$ into PCA space $\mathbf{R}$ of $\{\mathbf{x'}\}$, obtaining rotated dictionary $\mathbf{C}'_m:\{\mathbf{c}'_m(k)=\mathbf{R}\mathbf{c}_m(k), k=1\cdots K\}$ and rotated intermediate dataset $\{\mathbf{x^r}=\mathbf{R}\mathbf{x'}\}$.
\item Optimize $\mathbf{C}'_m$ by performing K-means on $\{\mathbf{x^r}\}$, initialized with $\mathbf{C}'_m$, using only the top $d_1$ dimensional data, then on the top $d_2$, next on $d_3, \cdots, and finally on d_I=d$ dimensions.
\item Rotate $\mathbf{C}'_m$ back to finish the optimization: $\mathbf{C}_m=\{\mathbf{R}^T\mathbf{c}'_m(k), k= 1\cdots K\}$
\end{enumerate}

The choice of $d_1 \cdots d_I$ have minor effect on the optimization of a dictionary. We choice $d_i=d^{i/I}, I=10$ in our experiments.

\subsection{Multi-path Encoding}
\label{encoding}

To encode with DA dictionaries, we seek the code that minimizes the quantization error $E$ for an input vector $\mathbf{x}$:
\begin{equation}
\label{app}
\begin{split}
E&=\lVert \mathbf{x}-\sum_{m=1}^M \mathbf{c}_m(i_m(\mathbf{x}))\rVert^2 \\ &=\sum_{m=1}^M\lVert \mathbf{x}-\mathbf{c}_m(i_m(\mathbf{x}))\rVert^2-(m-1)\lVert \mathbf{x}\rVert^2 \\
&\quad+\sum_{a=1}^M\sum_{b=1,b\neq a}^M  {\mathbf{c}_a(i_a(\mathbf{x}))}^\mathrm{T}\mathbf{c}_b(i_b(\mathbf{x}))
\end{split}
\end{equation}

The above algorithm is a typical fully connected MRF problem. Though the optimization of $E$ can be solved approximately by various existing algorithms, they're very time consuming\cite{babenko2015tree}. 

\newcommand\Xtilde{\stackrel{\sim}{\smash{\mathbf{x}}\rule{0pt}{1.1ex}}}
Similar to the concept of Locally Aggregating Encoding, if given an oracle the correct first $m-1$ encodings, can we effectively tell the correct encoding on the $m$-th part? Denote the correct encoding of a input vector as $\mathbf{x}\approxeq\mathbf{c}_1(i_1)+\mathbf{c}_2(i_2)+\cdots+\mathbf{c}_m(i_m)$, and the known $m-1$ correct encodings ${i_1, i_2,\cdots, i_{m-1}}$, we consider quantization error $E$ as a function of $\mathbf{c}_m(i_m)$:

\begin{equation}
\begin{split}\label{equ2}
E= & \lVert\mathbf{x}-(\hat{\mathbf{x}}+\mathbf{c}_m(i_m)+\Xtilde)\rVert^2\\
 = & \lVert\mathbf{x}-\hat{\mathbf{x}}\rVert^2 + \lVert\mathbf{x}-\Xtilde\rVert^2 + 2\hat{\mathbf{x}}^T\Xtilde\\
   & + \lVert\mathbf{x}-\mathbf{c}_m(i_m)\rVert^2 + 2\hat{\mathbf{x}}^T\mathbf{c}_m(i_m)+2\mathbf{c}_m(i_m))^T\Xtilde \\
   &-2\lVert\mathbf{x}\rVert^2
\end{split}
\end{equation}

where $\hat{\mathbf{x}}=\mathbf{c}_1(i_1)+\cdots+\mathbf{c}_{m-1}(i_{m-1})$, and
$\Xtilde=\mathbf{c}_{m+1}(i_{m+1})+\cdots+\mathbf{c}_{M}(i_{M})$

We seek the best $i_m$ among $1\cdots K$ to minimize $E$. In Equation \ref{equ2}, terms 1/2/3/7 are constant and negligible, terms 4/5 can be computed. Only the 6-th term cannot be computed because we don't know $\Xtilde$. We want it to be small so it won't seriously affects the final outcome. 

Thus we rearrange the dictionaries in the descending order of dictionary's elements variance. Note that DA learned from scratch naturally produces variance descending dictionaries. We further adopt beam search to encode a vector. That is, we maintain a list of best $L$ approximations of $\mathbf{x}$ on the first $(m-1)$ dictionaries: $\{\mathbf{a}_1^{m-1},\mathbf{a}_2^{m-1}, \cdots , \mathbf{a}_l^{m-1}\}$. Then we encode with the next dictionary $\mathbf{C}_m=\{\mathbf{c}_{m}(1), \mathbf{c}_{m}(2),\cdots,\mathbf{c}_{m}(K)\}$. We find $L$ combinations from $\{\mathbf{a}_{l}^{m-1}+\mathbf{c}_{m}(k)\}, l\in 1\cdots L, k\in 1\cdots K$ by minimizing the following objective function: 

\begin{equation}
\begin{split} \label{equ3}
\lVert \mathbf{x}-\mathbf{a}_{l}^{m-1}-\mathbf{c}_{m}(k)\rVert^2=&\lVert\mathbf{x}-\mathbf{a}_{l}^{m-1}\rVert^2+\lVert\mathbf{x}-\mathbf{c}_{m}(k)\rVert^2 \\
&-\lVert \mathbf{x} \rVert^2 + 2\mathbf{c}_{m}(k)^T\mathbf{a}_{l}^{m-1}
\end{split}
\end{equation}

We enumerate $KL$ combinations and select top $L$ candidates. For each combination in Equation \ref{equ3}: 
\begin{itemize}
\item The first term has been computed at the previous encoding step - one table lookup.
\item The second term $\lVert\mathbf{x}-\mathbf{c}_{m}(k)\rVert^2$ is pre-computed for each encoding vector taking $O(dK)$ time - one table lookup
\item The third term $\lVert \mathbf{x} \rVert^2$ is a negligible constant.
\item The last term involves $m$ table lookups and addition, with the inner-product of all dictionaries elements precomputed before the beam search procedure.
\end{itemize}    

To sum up, the time complexity is O($dK+mKL+KL\log L$) for encoding with one single dictionary. Note for fresh start DA, we don't need to encode the previously learned dictionaries excessively after we optimized a "zero" dictionary(i.e. learned a new dictionary). We report the $L$-distortion curve in Figure \ref{fgC}, we found that a relatively low $L=10$ could already achieve satisfactory encoding quality. We use this configuration in the rest of the experiments.

\subsection{Online Dictionary Learning}
\begin{figure*}[t]
    \subfigure[Online Learning]{
    	\includegraphics[width=0.34\linewidth]{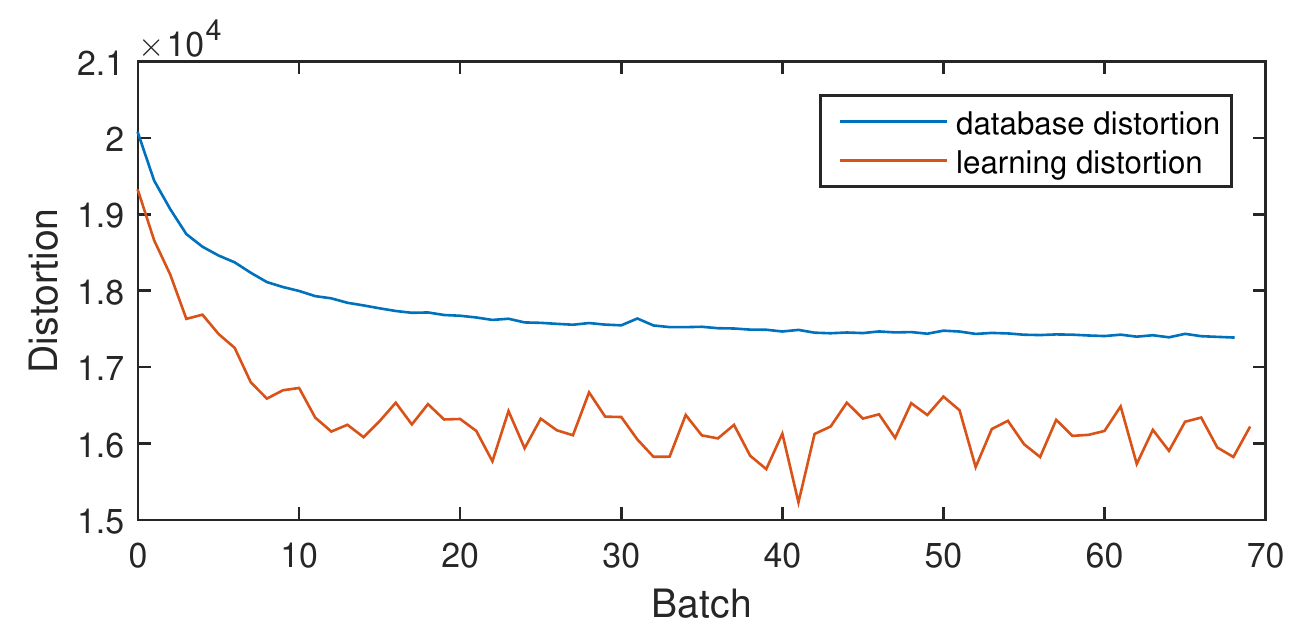}
    	    \label{fgA}
    }
    ~
    \subfigure[Training Time vs Distortion]{
         \includegraphics[width=0.34\linewidth]{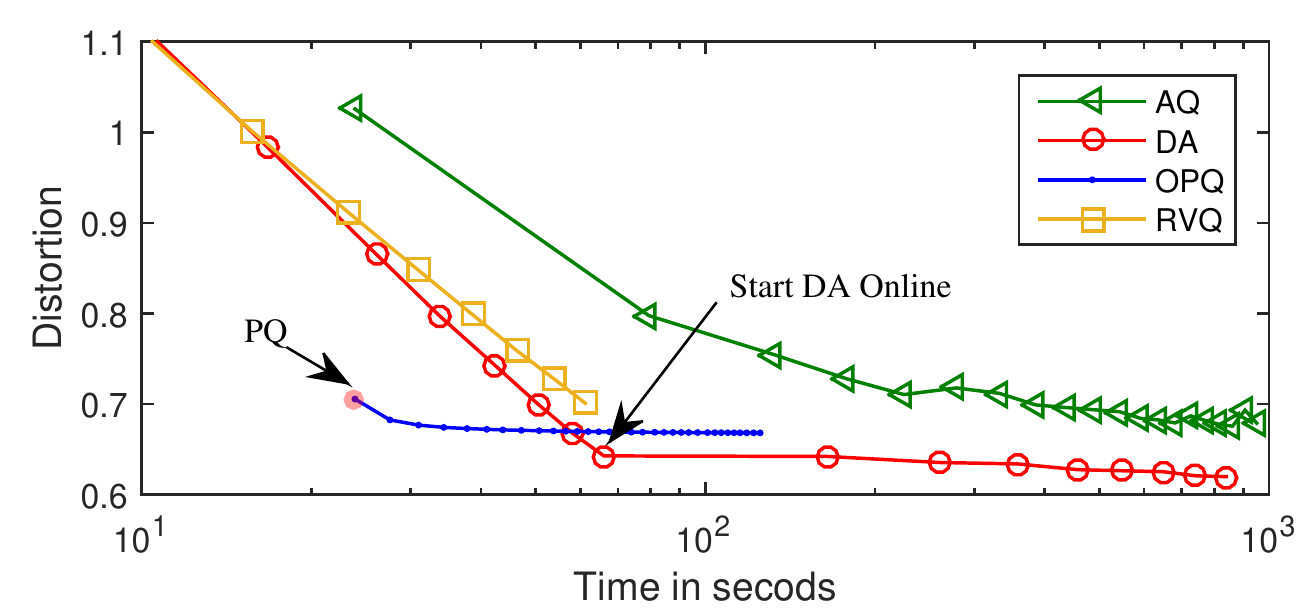}
         \label{fgB}
    }
    ~
    \subfigure[Multi-path Encoding]{
         \includegraphics[width=0.22\linewidth]{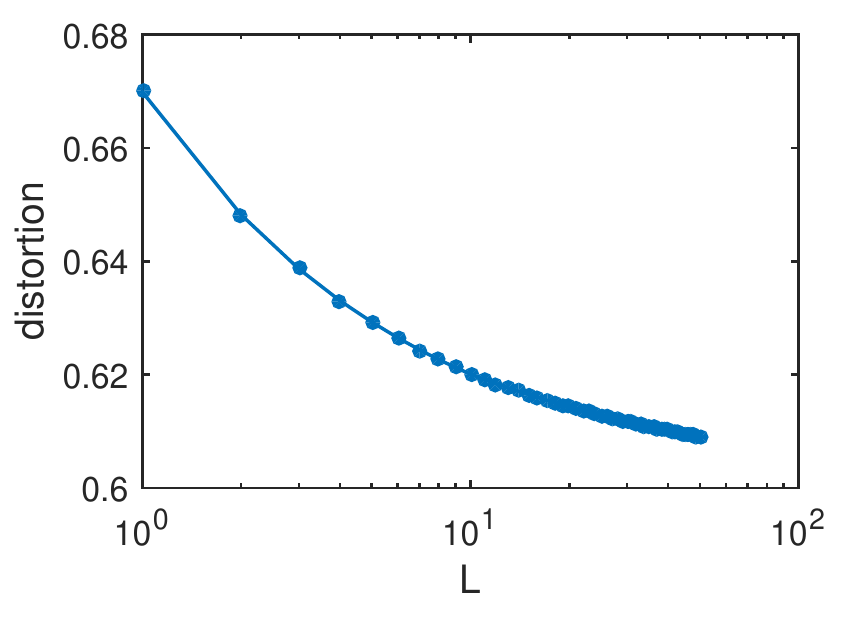}
         \label{fgC}
    }
    
    \caption{Empirical Analysis on Dictionary Annealing. \ref{fgA} \textit{Online learning with Dictionary Annealing} further lowers quantization error. Experiment on SIFT1B with $K=256, M=8$, feeding DA with 100,000 new vectors in batch. \ref{fgB}\textit{Training time vs distortion} on GIST1M: Whenever we obtained usable dictionaries for encoding, we encode the dataset and check time. Note: OPQ is initialized with PQ and DA continues optimization with online learning. Experiment runs on one GTX980 GPU. \ref{fgC} Encoding GIST1M dataset with Multi-path Encoding: we use DA-online dictionaries to examine how $L$ affects distortion } 
	\label{figOnline}
\end{figure*}

DA can be easily extended to an online learning mechanism to utilize even larger scale dataset, where clustering on all data could be prohibitive, or new data is not yet available currently. Online learning with DA can be done simply by optimizing the learned dictionaries to fit the new coming data. We report online learning result for SIFT1B\cite{jegou2011searching} dataset in Figure \ref{fgA}.

\section{Aggregating Tree}
\begin{figure*}[]
    \centering
    \includegraphics[width=1.0\linewidth]{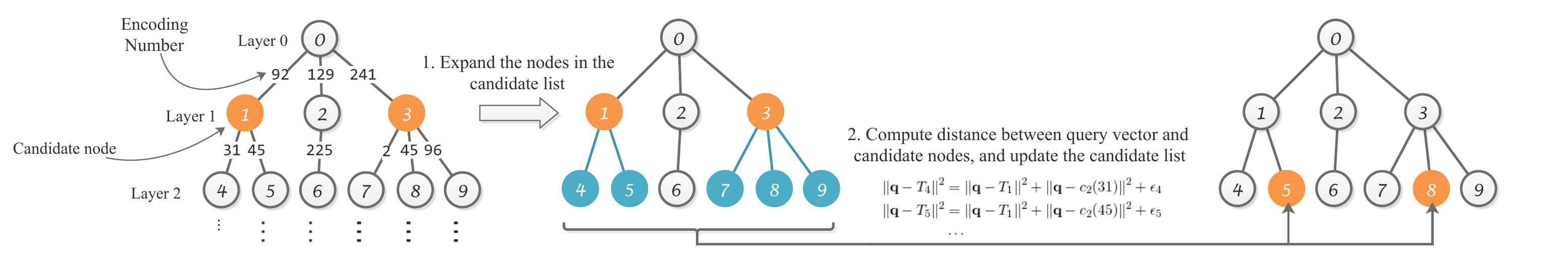}
    \caption{A toy demonstration of searching with A-Tree. The orange nodes are the current candidate list. The blue nodes are ready to enter the candidate list. The nodes closet to the query vector is selected in the new candidate list.}    
	\label{figIllustrationATree}
\end{figure*}

\begin{figure}[]
    \centering
    \includegraphics[width=1.0\linewidth]{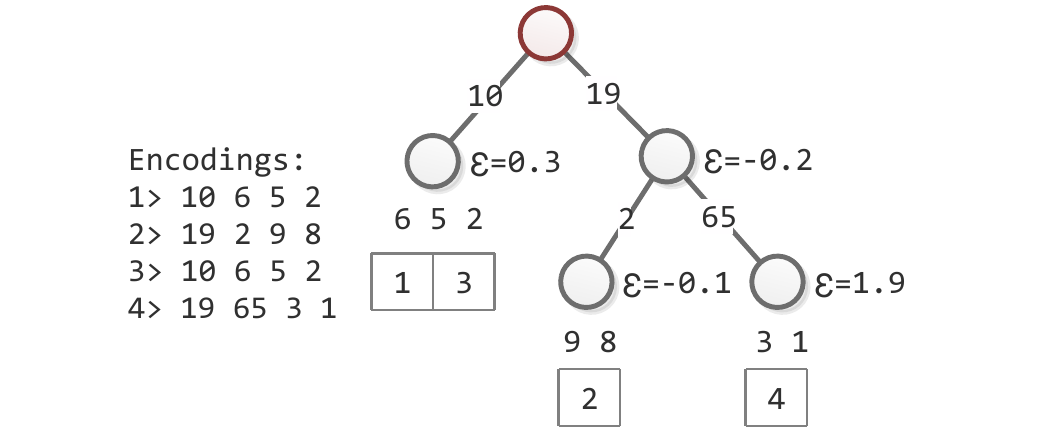}
    \caption{A toy illustration of the structure of A-Tree} 
	\label{figIllustrationATreeStructure}
\end{figure}

We are now able to adopt bound-and-branch methodology to high-dimensional data by Aggregating Tree(A-Tree). After obtaining HCLAEs with DA, A-Tree is constructed according to the encodings like a radix tree(each node that is the only child is merged with its parent), except that we only merge leaf nodes. A-Tree effectively presents the quantized dataset, with all encodings written directly on the tree. A demonstrative structure of A-Tree is shown in Figure \ref{figIllustrationATreeStructure}.

To perform non-exhaustive search on A-Tree, the idea is to maintain a candidate list like in multi-path encoding. First we determine a candidate list with size limit for each layer as: $L_1, \cdots, L_M$. Given a query vector $\mathbf{q}$, we start with an initial candidate list containing only the root node, and iteratively do the following for $M$ times (The procedure is illustrated in Figure \ref{figIllustrationATree}):
\begin{enumerate}
\item Replace the nodes in candidate list with their children. If the node has no children, it stays in the candidate list.
\item If the size of the candidate list exceeds $L_i$($i$ is the current iteration number), shrink it to $L_i$, and discard the nodes distant to the query vector.
\end{enumerate}

We have to record some extra information on each node to allow fast distance computation. Let $m$ denote the depth of a node $T$, and $p_1, \cdots, p_m$ is the path from the root to this node, we record:
$$\epsilon=\sum_{i=1}^{m-1} {\mathbf{c}_m(p_m)}^\mathrm{T}\mathbf{c}_i(p_i)$$
for $T$. When we compute the distance between $\mathbf{q}$ and $T$ (reconstructed as $\sum_{i=1}^m \mathbf{c}_i(p_i)$), we have known the distance between $\mathbf{q}$ and $T$'s father $T'$, we have:

\begin{multline}
\centering
\begin{split}
\lVert\mathbf{q}-T\lVert^2&=\lVert\mathbf{q}-T'\lVert^2+\lVert\mathbf{q}-\mathbf{c}_m\lVert^2+{\mathbf{c}_m(p_m)}^\mathrm{T}T' \\
&\quad-\lVert\mathbf{q}\rVert^2\\
{\mathbf{c}_m(p_m)}^\mathrm{T}T'&={\mathbf{c}_m(p_m)}^\mathrm{T}\sum_{i=1}^{m-1}\mathbf{c}_i(p_i)=\epsilon
\end{split}
\end{multline} 

Thus the distance computation between a node and the query can be done efficiently in $O(1)$. \footnote{We can further reduce the number of additions and table look ups with a smart implementation, please refer to the supplementary material.}

After the above steps we have obtained the list of approximate nearest neighbors. The configuration of $L_i$ has an influence on the final search quality, which will be discussed in the Experiments Section. Candidate listing with A-Tree is highly efficient: the overall time complexity is $O(N'+dKM)$, where $N'$ refers to the total number of nodes traversed. Note that A-Tree is a tree structure, the performance is heavily dependent on the implementation.

\section{Experiments}

\begin{figure*}[]
    \subfigure[Search quality]{
    	\includegraphics[width=0.30\linewidth]{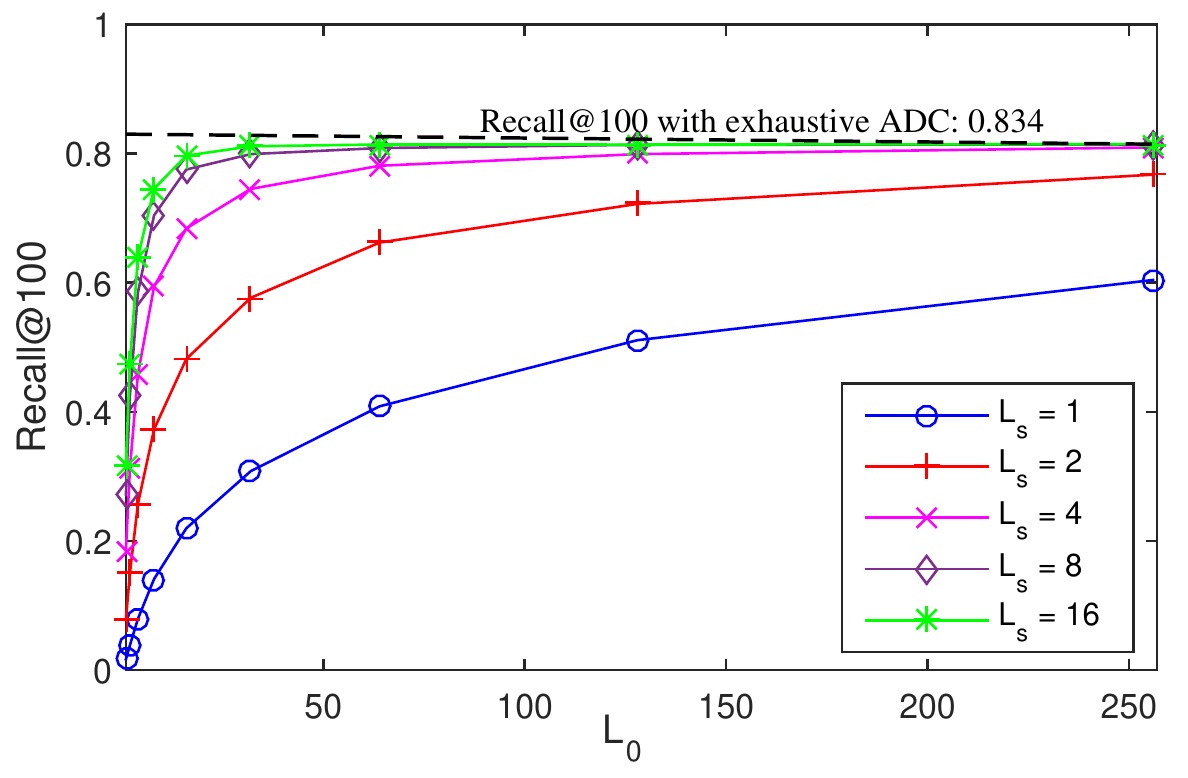}
    	\label{sq}
    }
    \subfigure[Number of nodes visited per query]{
        \includegraphics[width=0.3\linewidth]{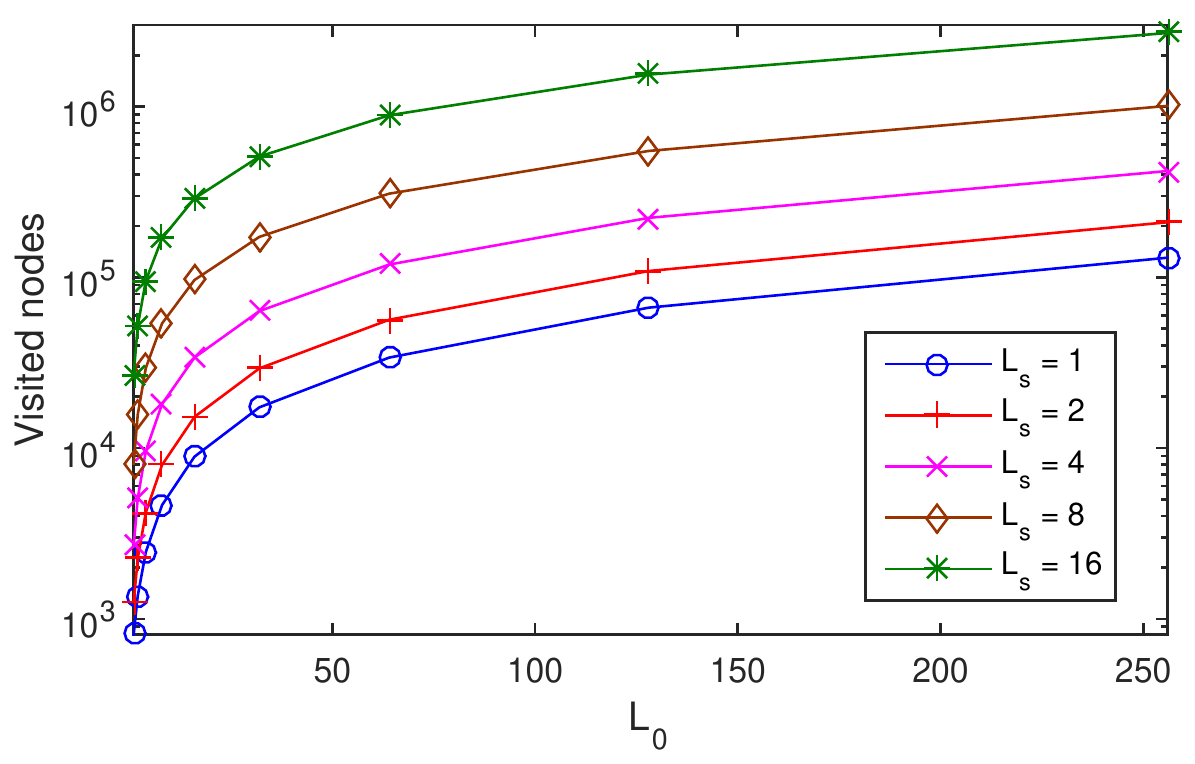}
        \label{nvpq}
    }
    \subfigure[Performance curve]{
        \includegraphics[width=0.3\linewidth]{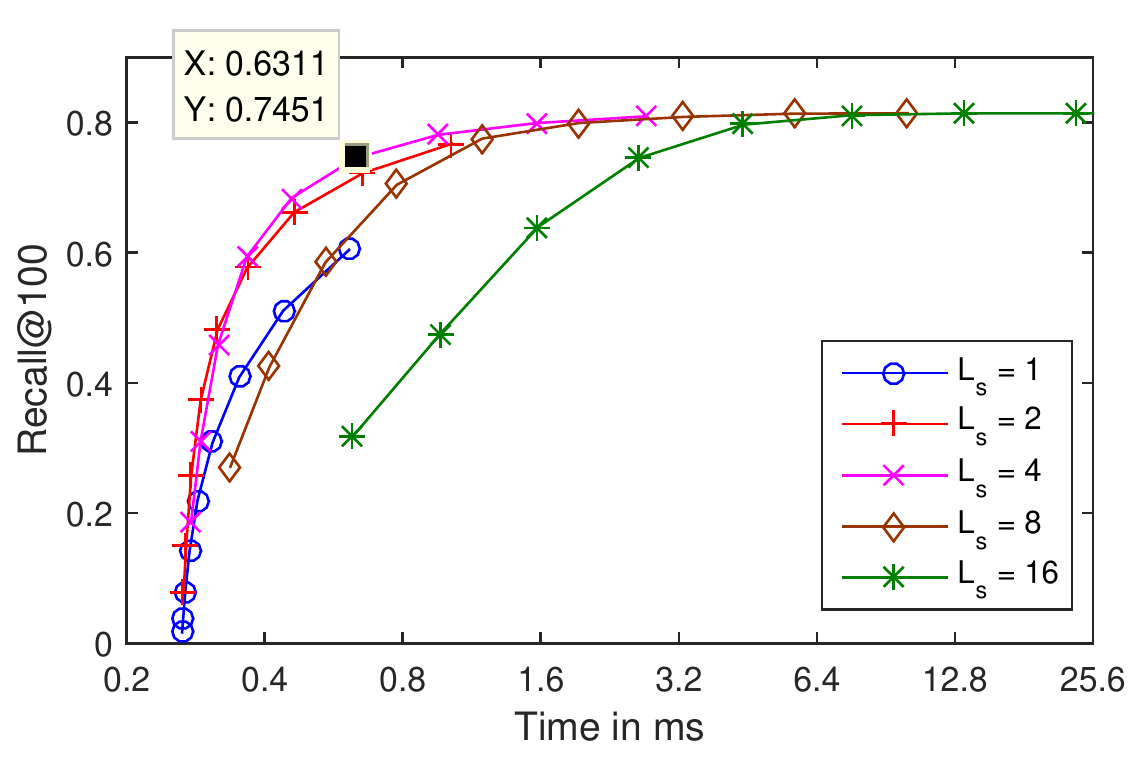}
        \label{pef}
    }
    \caption{Non-exhaustive search on A-tree, candidate list size for each layer is $L_i=L_0L_s^i$, where $L_0=1,2,4,\cdots,256$. On the last layer we shrink the list to 100 and check if the true nearest neighbor is in the list.} 
	\label{figL0Ls}
\end{figure*}
In this section we present the experimental evaluation of Dictionary Annealing and A-Tree. All experiments are done on an Core i7 running at 3.5GHz with 16G memory, single threaded.

\subsection{Datasets}
We use the following datasets commonly used to validate the efficiency of ANN methods: SIFT1M\cite{pq}, contains one million 128-d SIFT \cite{sift} features and 10000 queries; GIST1M\cite{pq}, contains one million 960-d GIST \cite{gist} global descriptors and 1000 queries; SIFT1B\cite{jegou2011searching} contains one billion 128-d SIFT feature as base vectors, 10K queries.

\subsection{Performance of Dictionary Annealing}
We compare the following state-of-the-art encodings: Optimized Product Quantization(OPQ), Composite Quantization(CQ)\cite{composite}, Additive Quantization(AQ)\cite{babenko2014additive}, Tree Quantization(TQ) and it's optimized version Optimized Tree Quantization\cite{babenko2015tree}. We re-implemented AQ and OPQ by   ourselves, and reproduce the results from \cite{composite} and \cite{babenko2015tree} to present the evaluation. We choose the commonly used configuration: $K=256$ as the dictionary size and $M=8,16$ for all methods.

\renewcommand{\arraystretch}{1.2}
\begin{table}
\centering
\tiny
    \begin{tabularx}{0.926\linewidth}{|c||c|c|c|c|c}
    	\hline
    	 Methods   &        8B-SIFT1M        &       16B-SIFT1M       & 8B-GIST1M & 16B-GIST1M \\ \hline\hline
    	    AQ     &        19196.26         &        9799.86         &  0.6785   &   0.5277   \\ \hline
    	   OPQ     &        22239.78         &        10468.39        &  0.6973   &   0.5361   \\ \hline
    	    PQ     &        23540.75         &        10534.82        &  0.7056   &   0.6976   \\ \hline
    	    TQ     & (about\textasciitilde 20000) & (about\textasciitilde 9000) &     -     &     -      \\ \hline
    	offline-DA &        \textbf{18416.55}         &        \textbf{9444.11}         &  \textbf{0.6456}   &   \textbf{0.4847}   \\ \hline
    	online-DA  &        \textbf{16573.20}         &        \textbf{5901.43}         &  \textbf{0.6201}   &   \textbf{0.4583}   \\ \hline
    \end{tabularx}
    \caption{A comparison of quantization error between different quantization methods. We used $K=256, M=8/16$ for all methods. Values in brackets are from \cite{babenko2015tree}.}
        \label{distortion}
\end{table}

\begin{figure}[]
    \centering

    \subfigure[Performance on SIFT1M]{
    	\includegraphics[width=0.47\linewidth]{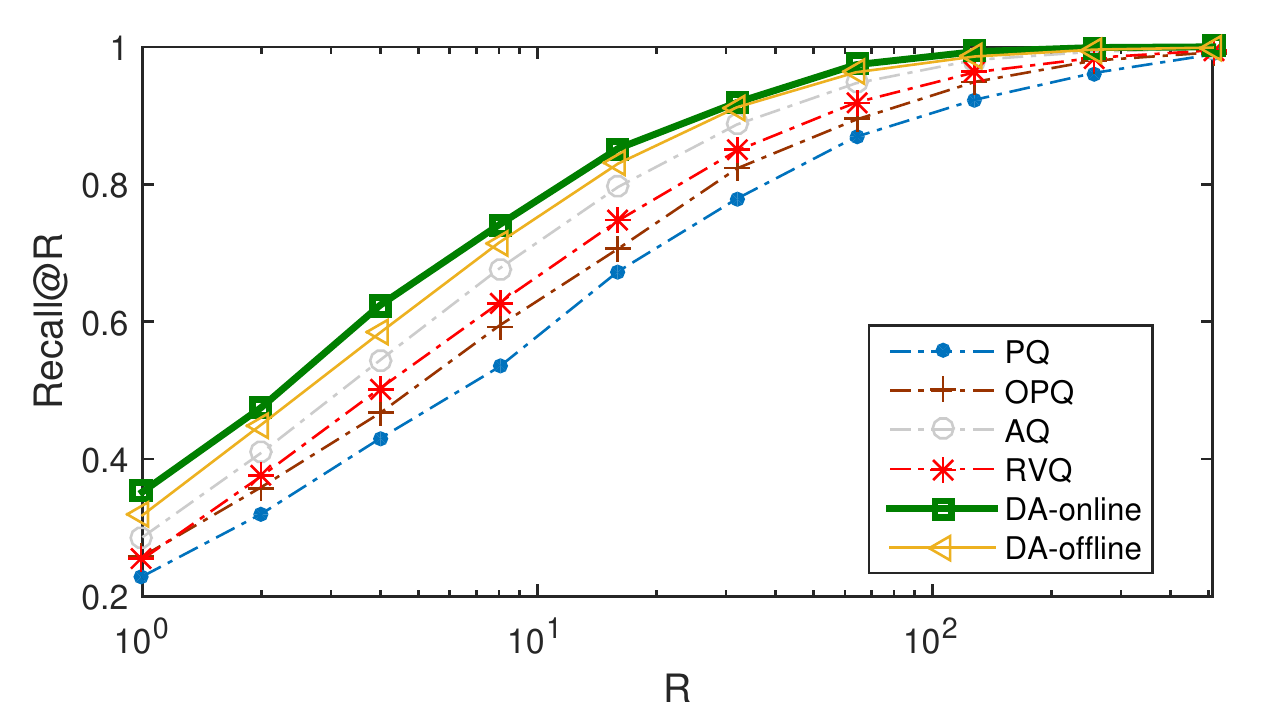}
    }
    ~
    \subfigure[Performance on GIST1M]{
    	\includegraphics[width=0.455\linewidth]{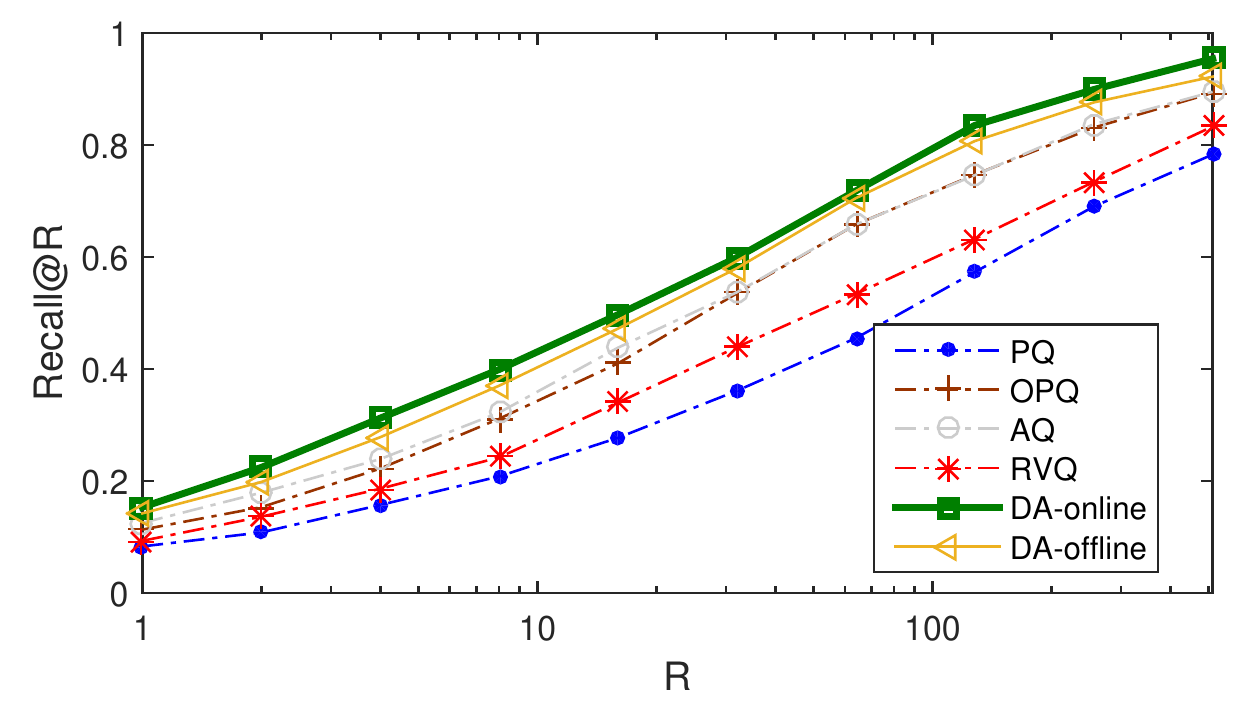}
    }
    \caption{The comparison of Recall vs Number of items retrieved curve with 64bits encodings.} 
	\label{figRecall}
\end{figure}

We use SIFT1M and GIST1M for evaluation, and train all methods on the training set and encode the whole dataset. We also train online DA with all the data\footnote{We didn't train other methods on the whole dataset because they require too much memory}, and report the training time vs distortion graph to in Figure \ref{fgB}, DA runs almost as fast as RVQ and much faster than AQ. The quantization error is presented in Table \ref{distortion}, our AQ has a much lower quantization error than other state-of-the-art. We perform exhaustive NN-search and report the performance of different methods in Figure \ref{figRecall}. It can be seen that DA consistently perform better than other state-of-the-art methods. It's online learning version further pushes the performance of the encodings higher, for example by \textbf{13.6\%} lower distortion and \textbf{23.07\%} higher recall@1 for NN-Search on 8-Bytes SIFT1M encoding. 

\subsection{Searching with Aggregating Tree}

Now we evaluate the performance of Aggregating Tree. We constructed an A-Tree for SIFT1B(DA-online with 10M vectors of the dataset, $M=8, K=256$). We design the A-Tree to be computation efficient\footnote{Implementation details are presented in supplementary materials}. The outcome data-structure occupies 14.53GB (total 1,224,574,028 Nodes consisting of 988,853,094 leaf nodes and 235,720,934 internal nodes) memory for SIFT1B with 64-bit encoding, including vectors ID.

\renewcommand{\arraystretch}{1.2}
\begin{table}
\tiny
\centering
    \begin{tabularx}{0.8\linewidth}{|c||X|X|c|}
    	\hline
    	  System    & Recall@1                            & Recall@100                 &      Query Time       \\ \hline
    	  IVFADC    & (\textit{0.088})0.107               & (\textit{0.733})0.729      & (\textit{74ms}) 65ms   \\ \hline
    	Multi-D-ADC & (\textit{0.158})0.149               & (\textit{0.706})0.717      & (\textit{6ms}) 3.4ms   \\ \hline
    	 Multi-ADC  & (\textit{\textasciitilde0.05})0.064 & (\textasciitilde 0.6)0.582 &        3.2ms           \\ \hline
    	   LOPQ     & (\textit{0.199})0.182               & (\textit{0.909}) 0.890     &         69ms           \\ \hline
    	  A-Tree    & 0.137                               & 0.7451                     &        0.63ms          \\ \hline
    \end{tabularx}
    \caption{Comparison of various non-exhaustive search methods, for Multi-D-ADC, Multi-ADC, $K=2^14$ and $T=10000$. Result in brackets are taken from \cite{babenko2012inverted} and \cite{kalantidis2014locally}.}
        \label{recall}
\end{table}

The choice of $L_i$ is important for searching with A-Tree. The encodings by DA don't always guarantee the local vectors have the exact same prefix. We let $L_i=L_0L_s^i$ in our experiments. Figure \ref{nvpq} reports the number of nodes traversed, though $L_i$ grows exponentially, the total number of traversed nodes is limited. We also report the performance of an exhaustive ADC(\textit{7.2s per query}) on the whole dataset. A-Tree delivers asymptotic performance to exhaustive ADC by magnitudes of acceleration as shown on Figure \ref{sq}. One can use a longer encoding for preciser search result. We finally draw the performance curve of A-Tree in Figure \ref{pef}. A-Tree achieves an amazing speed at \textbf{0.63ms} with a high search quality of \textbf{74.51\%} Recall@100, at the elbow of the curve.

In Table \ref{recall} we compared A-Tree with our speed optimized implementations of IVFADC\cite{pq}, Locally Optimized Product Quantization\cite{kalantidis2014locally}, Multi-D-ADC and Multi-ADC\cite{babenko2012inverted}. A-tree achieves \textbf{9.5x} acceleration over Multi-D-ADC and over \textbf{117x} accleration over IVFADC with comparable performance. We think this is mainly because:
\begin{enumerate}
\item A-Tree joins candidate listing and re-ranking procedures together to avoid excessive "pre-computation". It also make A-Tree cache friendly. While other methods requires many times of re-calculating the look-up table and cache unfriendly. 
\item A-Tree is based on HCLAE so a shorter list of candidates could already achieve satisfying result. While for IVFADC, a typical length of candidates is 80M on SIFT1B dataset.
\item DA produces high quality encoded dataset, especially with online learning(Recall@100:\textbf{0.834} on 64 bit, compared to Composite Quantization \cite{composite} :\textasciitilde0.7, OPQ: \textasciitilde0.65, PQ: \textasciitilde0.55)
\end{enumerate}

\section{Conclusion}
In this paper, we introduced the concept of High Capacity Locally Aggregating Encodings(HCLAE) for ANN search. We proposed Dictionary Annealing to produce HCLAE, and Aggregating Tree to perform fast non-exhaustive search. Empirical results on datasets commonly used for evaluating ANN search methods demonstrated our proposed approach significantly outperforms existing methods.

\bibliographystyle{aaai}
\bibliography{sigproc}  % sigproc.bib is the name of the Bibliography in this case

\end{document}